\documentclass{article}
\usepackage{arxiv}

\usepackage{amsmath,amsfonts,bm}

\def\eqref#1{equation~\ref{#1}}

\def\1{\bm{1}}

\def\vx{{\bm{x}}}

\DeclareMathAlphabet{\mathsfit}{\encodingdefault}{\sfdefault}{m}{sl}
\SetMathAlphabet{\mathsfit}{bold}{\encodingdefault}{\sfdefault}{bx}{n}

\def\gL{{\mathcal{L}}}

\usepackage{url}
\usepackage{booktabs}
\usepackage{amsmath}
\usepackage{amssymb}
\usepackage{graphicx}
\usepackage{multirow}
\usepackage{xcolor}
\usepackage{subcaption}
\usepackage{enumitem}
\usepackage{float}
\usepackage{natbib}
\usepackage{hyperref}
\hypersetup{breaklinks=true, hidelinks}

\setcitestyle{authoryear,round,citesep={;},aysep={,},yysep={;}}

\renewcommand{\headrulewidth}{0.4pt}
\makeatletter
\renewcommand{\maketitle}{%
  \par
  \begingroup
    \renewcommand{\thefootnote}{\fnsymbol{footnote}}
    \long\def\@makefntext##1{%
      \parindent 1em\noindent
      \hbox to 1.8em{\hss $\m@th ^{\@thefnmark}$}##1
    }%
    \@maketitle
    \@thanks
  \endgroup
  \let\maketitle\relax
  \let\thanks\relax
}
\renewcommand{\@maketitle}{%
  \vbox{%
    \hsize\textwidth
    \linewidth\hsize
    \vskip 0.1in
    \@toptitlebar
    \centering
    {\LARGE\sc \@title\par}
    \@bottomtitlebar
    \vskip 0.1in
    \def\And{%
      \end{tabular}\hfil\linebreak[0]\hfil%
      \begin{tabular}[t]{c}\bf\rule{\z@}{24\p@}\ignorespaces%
    }%
    \def\AND{%
      \end{tabular}\hfil\linebreak[4]\hfil%
      \begin{tabular}[t]{c}\bf\rule{\z@}{24\p@}\ignorespaces%
    }%
    \begin{tabular}[t]{c}\bf\rule{\z@}{24\p@}\@author\end{tabular}%
  \vskip 0.4in \@minus 0.1in \center{\@date}   \vskip 0.2in
  }%
}
\makeatother

\makeatletter
\renewcommand\section{\@startsection{section}{1}{\z@}%
  {-2.0ex \@plus -0.5ex \@minus -0.2ex}%
  { 1.5ex \@plus  0.3ex \@minus  0.2ex}%
  {\large\sc\raggedright}}
\renewcommand\subsection{\@startsection{subsection}{2}{\z@}%
  {-1.8ex \@plus -0.5ex \@minus -0.2ex}%
  { 0.8ex \@plus  0.2ex}%
  {\normalsize\sc\raggedright}}
\renewcommand\subsubsection{\@startsection{subsubsection}{3}{\z@}%
  {-1.5ex \@plus -0.5ex \@minus -0.2ex}%
  { 0.5ex \@plus  0.2ex}%
  {\normalsize\sc\raggedright}}
\makeatother

\newcommand{\cctf}{teacher forcing}
\def\vepsilon{{\bm{\epsilon}}}

\title{Real-Time Joint Audio--Video Generation by Parallel Adapter Composition}

\author{
  Jingyu Li$^{1}$\thanks{Equal contribution.}\hspace{1.5em}
  Xiaoxiao Xiang$^{1}$\footnotemark[1]\hspace{1.5em}
  Yiwen Guo$^{2}$\thanks{Corresponding author.}\\
  \normalfont $^{1}$LIGHTSPEED \quad $^{2}$Independent Researcher
}
\date{}

\begin{document}

\maketitle

\begin{abstract}
Deploying a joint audio--video diffusion transformer for real-time, interactive
generation normally requires two essential modifications: block-autoregressive
attention, so frames can be emitted before the whole clip is finished, and few-step
sampling, so each block is cheap.
Conventionally, the streaming video literature obtains both capabilities
from a \emph{chained} pipeline. It first distills a bidirectional teacher
into a causal student, then into a few-step one, or proceeds in reverse order.
Each stage of such a chain fine-tunes the weights the previous one produced,
so a later objective can undo an earlier capability.
Following the idea of model merging, we show that on a packed audio--video
backbone the two capabilities can be
acquired \emph{in parallel}. A causal
adapter is trained against the \emph{frozen} backbone, and an off-the-shelf few-step
adapter provides the few-step capability. As the two edit different
functional axes, we predict, and then verify, that their weight-update
directions are near-orthogonal, without any explicit orthogonality
constraint during training. Orthogonal updates should combine without
interfering, so parallel composition is a direct sum. The two adapters are
simply \emph{added} at inference, with no joint training, yielding few-step,
streaming audio--video whose image quality tracks the bidirectional teacher. 
Compared to the chained baselines, the composed model
matches or beats them on most metrics, making parallel composition a
practical approach. The resulting streaming system generates joint audio--video in
real time, ${\approx}26$\,fps at $480{\times}832$ without quantization, and sustains $30$\,s of
continuous generation with stable image quality. Demo Page: \url{https://pac-demo-2027.github.io/demo/}
\end{abstract}

\section{Introduction}
\label{sec:intro}

Joint audio--video generation has matured quickly: recent models produce
synchronized audio and video at high fidelity
\citep{liu2026javisdit,hacohen2026ltx2,minimax2026h3,seedance2026}. Most of
them, however, are bidirectional: the frame count must be fixed before sampling,
nothing is emitted until the last step finishes, and every step costs
quadratically in the requested duration. Applications built on real-time
interaction \citep{bai2026mainecoon,huang2026wanstreamer} --- interactive
avatars, live agents --- call for the opposite properties: a bounded
time-to-first-frame, a per-chunk cost independent of session length, and no
length ceiling. 

Two capabilities are required. \emph{Causalization} partitions the clip into
chunks and denoises chunk $k$ from chunks $0{:}k{-}1$, so the outputs are
emitted incrementally and a chunk is played as soon as it is ready.
\emph{Few-step generation} shrinks the per-chunk cost from tens of denoising
steps to a handful, which is the key to real-time generation. Learning them
may take several training stages, and each stage may train a
single objective or both. CausVid and
Self-Forcing \citep{yin2025causvid,huang2025selfforcing} learn both under one
distillation objective, training a causal few-step student directly from the
bidirectional teacher.
Causal Forcing and its successor \citep{zhu2026causalforcing,zhao2026causalforcingpp} split the
process, training a multi-step autoregressive model first and distilling it to
few steps afterwards; OmniForcing \citep{su2026omniforcing} runs the reverse
order, converting a distilled bidirectional few-step generator into an autoregressive one. The split
has its appeal: each stage optimizes a single objective rather than
mixing several, can start from an existing checkpoint, and can be analyzed in
isolation. But the training stages interact: whichever runs first changes the
base the second trains on. We refer to the sequential recipes as \emph{chained}
in what follows.

\begin{figure}[t]
\centering
\includegraphics[width=\linewidth]{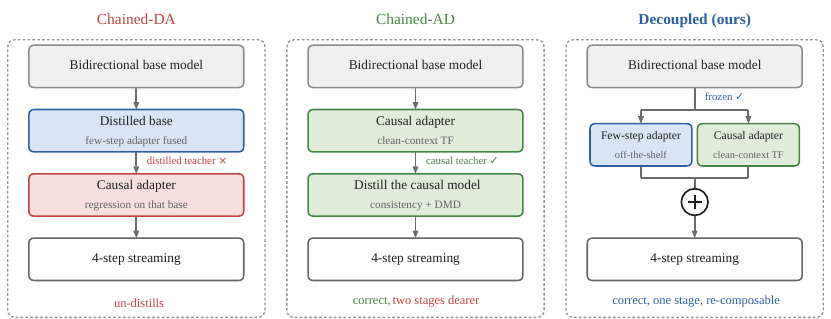}
\caption{\textbf{Three routes to a few-step streaming model, and what the order
costs.} Distilling last keeps the chain viable, at two extra stages; the reverse
order, trained with a plain regression loss, un-distills. The mechanisms are in
\S\ref{sec:routes}.}
\label{fig:routes}
\end{figure}

We follow the idea of splitting the learning of causalization and few-step
distillation, and ask whether the chain is necessary to achieve the desired
capabilities. The experimental results demonstrate that it is not: the two
capabilities can be acquired \emph{in parallel}, by weight addition at inference. The
principle is model merging / task arithmetic
\citep{ilharco2023taskarithmetic}: updates fine-tuned independently against
one frozen base can be added in weight space, in the small-edit regime, while
each keeps its function. Our route does exactly this. The causal adapter $\Delta_c$ is trained against
the frozen $W$ with clean-context teacher forcing---a plain flow-matching
regression, and the only objective we train. The few-step adapter
$\Delta_d$ is off the shelf, distilled under \emph{bidirectional} attention
where distillation is already solved. The two are then \emph{added} at
inference, with no joint or chained training, so neither stage ever sees a base
the other has altered. Two chained baselines are included for comparison: \textsc{Chained-DA}
(distill, then causalize) retrains the causal adapter on the distilled weights
with the plain flow-matching regression, the same order as OmniForcing
\citep{su2026omniforcing}.
\textsc{Chained-AD} (causalize, then distill) is the causal-forcing recipe
\citep{zhu2026causalforcing,zhao2026causalforcingpp}. The differences between the three approaches
can be seen in Figure~\ref{fig:routes}, and our recipe is denoted as \textsc{Decoupled}.

Our contributions are threefold. \textbf{(1) Parallel composition works:} the
two adapters, trained
independently against the one frozen backbone and simply added at inference,
match the teacher on appearance and the single-frame distribution, and beat
\textsc{Chained-DA} on most held-out
metrics, while trailing \textsc{Chained-AD} on lip sync. There is no
chained or joint training, none of the chains' extra stages, and the causal
adapter stays reusable across step counts and on both the undistilled
and the distilled weights
(\S\ref{sec:distfree}).
\textbf{(2) A real-time streaming system from short-clip training:} the
composed model generates joint audio--video at ${\approx}26$\,fps at
$480{\times}832$ on six GPUs without quantization, and a causal adapter
trained on $5$--$10$\,s clips extrapolates to stable $30$\,s generation, where
\textsc{Chained-DA} decays and falls silent
(\S\ref{sec:causalcost}, Table~\ref{tab:longrun-windows}).
\textbf{(3) A predicted, then verified, geometry:} the two updates read
overlapping input subspaces but write into near-orthogonal directions, so
their sum needs no constraint, and weight cancellation is ruled out as
\textsc{Chained-DA}'s failure mode (\S\ref{sec:geom}).
All comparisons share one pipeline and reference the teacher's own generation
(Table~\ref{tab:main}); the ordering carries over to three public benchmarks
scored against real sources (\S\ref{sec:external}, Appendix~\ref{app:public}).

\section{Related Work}
\label{sec:related}

\paragraph{Joint audio--video generation.}
Video generation has been studied for years, but the focus has been mostly on
the visual modality alone
\citep{yang2024cogvideox,kong2024hunyuan,wan2025video}, with no accompanying
audio. Audio-driven portrait animation
\citep{prajwal2020wav2lip,cui2024hallo3,gan2025omniavatar} couples the two,
generating video conditioned on a driving audio track; the coupling is one-way,
the video following the audio, which itself cannot be freely prompted.
Generating \emph{both} jointly is developing rapidly
\citep{liu2026javisdit,hacohen2026ltx2,minimax2026h3,cosmos3}, the two
modalities produced simultaneously by one model. LTX-2
\citep{hacohen2026ltx2} uses a two-tower structure, one tower per modality,
with cross-attention sharing knowledge between the towers to keep the two
synchronized. In Cosmos 3 and MiniMax-H3, audio and video tokens are arranged
in a single sequence processed by one DiT.

\paragraph{Autoregressive and streaming diffusion.}
Diffusion Forcing \citep{chen2024diffusionforcing} and Rolling Diffusion
\citep{ruhe2024rolling} apply per-position noise levels, so early positions
with less noise are generated first. A family of later methods distills a
bidirectional teacher into a causal few-step student. CausVid
\citep{yin2025causvid} initializes the causal student on the teacher's ODE
pairs, then refines it with asymmetric DMD. Self-Forcing
\citep{huang2025selfforcing} additionally trains on the student's own
rollouts to close the train--test gap. Self-Forcing++
\citep{cui2025selfforcingpp} scales AR video generation to minute-long videos.
OmniForcing \citep{su2026omniforcing} is the closest work: it brings the
same line to joint audio--video, but as a multi-stage chain that first
distills a bidirectional few-step generator and then converts it to
autoregressive, the order our \textsc{Chained-DA} instantiates in plain
form. The recipe is involved, and that order is the weakest four-step
system in Table~\ref{tab:main}. Causal Forcing \citep{zhu2026causalforcing} and its successor
\citep{zhao2026causalforcingpp} run the stages in the reverse order, training
a multi-step autoregressive model first and distilling it last with
consistency and distribution-matching objectives on the student's own
rollouts; this is the recipe our \textsc{Chained-AD} follows. Others
scale or extend chunkwise autoregression
\citep{kim2024fifo,henschel2024streamingt2v,chen2025skyreels,magi2025}; all
are video-only except OmniForcing.

\paragraph{Model merging and adapter composition.}
Model merging, surveyed by \citet{yang2026mergingsurvey}, combines
independently fine-tuned models by arithmetic on their weights, e.g.\ task
arithmetic \citep{ilharco2023taskarithmetic}, TIES~\mbox{\citep{yadav2023ties}}
and DARE \citep{yu2024dare}, and rests on weight disentanglement
\citep{ortizjimenez2023tangent,jin2025attention}. For LoRA, OSRM
\citep{zhang2025osrm} \emph{imposes} orthogonality on the subspaces before
fine-tuning. In this work, two adapters are learned under
heterogeneous conditions (bidirectional distillation vs.\ block-causal
causalization) and edit different functional axes; they therefore tend to be
near-orthogonal and add directly. Avatar-Forever
\citep{avatarforever2026} likewise trains two branches in parallel and merges
them in weight space, but for audio-driven avatars, where the audio track is
given rather than generated, and without any analysis of when the sum is safe.

\section{Method}
\label{sec:method}
The backbone throughout is MiniMax-H3 in its first-frame-conditioned
audio--video (FL2VA) configuration, which runs $50$ bidirectional denoising
steps per generation. It is a diffusion transformer
trained by rectified flow \citep{liu2023rectified,lipman2023flowmatching} to
regress the instantaneous velocity ($t = 1-\sigma$; $t{=}1$ clean); its
architecture and VAE grid are given in
Appendix~\ref{app:geometry}. This undistilled model is the
\textsc{Teacher} of Table~\ref{tab:main}. The few-step adapter $\Delta_d$ is
off the shelf; the causal adapter $\Delta_c$ is trained against the frozen
backbone. Both are LoRAs, and they are added at inference.

\subsection{Limitations of the Chained Baselines}
\label{sec:routes}

In the chained routes, the training order of the adapters may result in different performance,
as the second adapter is trained on the weights of $W + \Delta_{\text{first}}$.

\textsc{Chained-DA} (distill, then causalize) fuses the public lightx2v Turbo
LoRA \citep{lightx2v2026turbo}, a third-party $4$-step distillation of $W$, into
the backbone, then trains a fresh causal adapter $\Delta_c'$ on the distilled
weights $W' = W + \Delta_d$ by teacher forcing (\S\ref{sec:cctf}) with the plain
flow-matching objective of Eq.~\ref{eq:loss}.

However, $\Delta_d$ predicts where a \emph{large} jump lands rather than the
instantaneous velocity the teacher regresses, so $W'$ no longer
minimizes the flow-matching objective; training $\Delta_c'$ with that objective
then pulls $\Delta_d$'s jump field back toward instantaneous velocity and
\emph{un-distills} it---four steps may no longer suffice.

\textsc{Chained-AD} (causalize, then distill) keeps the causal stage against the
frozen $W$ and distills the few-step property \emph{afterwards} on the causal
student itself, following the causal-forcing++ recipe
\citep{zhao2026causalforcingpp}: training on the student's own no-gradient
rollouts, with consistency targets along the teacher's Euler trajectory and a
DMD stage (Appendix~\ref{app:reverse}). The route is viable
(\S\ref{sec:causalcost}), but it comprises three stages and three objectives, two of
them entangled with the streaming cache, and the result is tied to the
weights it was trained on.

The two chains thus differ in both order and objective, and both share two
structural limitations. First, errors propagate along the chain: each adapter
trains on what the previous one produced, so an early weakness becomes part of
the next stage's training environment. The un-distillation of
\textsc{Chained-DA} is one instance; the anchor and online critic that
\textsc{Chained-AD} needs to keep its rollouts stable is another. Second, a
chained adapter cannot reuse existing components: the few-step capability of
\textsc{Chained-AD} must be distilled on the causal model, where nothing
off-the-shelf exists, and every chained adapter only works on the base it was
trained on (\S\ref{sec:flexible}). The parallel route removes both: each
capability trains against the same frozen $W$, so the two can be optimized
independently and either side can be an existing component. We take the
few-step adapter off the shelf and train only the causal one, a single plain
regression (\S\ref{sec:cctf}) with no ODE solver, EMA target, or critic. The
rest of this section develops it.

\subsection{Clean-Context Teacher Forcing}
\label{sec:cctf}

In streaming video generation, a video is split into chunks, and each chunk is generated autoregressively.
Chunk $k$ is denoised from noise, conditioned on the finished chunks before it,
\begin{equation}
\vx_k \sim p_{\Delta_c}\big(\,\vx_k \mid \bar{\vx}_0, \dots, \bar{\vx}_{k-1}\,\big),
\label{eq:ar}
\end{equation}
where $\bar{\vx}$ marks a chunk committed at full quality. In this work, we train $\Delta_c$
by teacher forcing: the clean history $[\vx_0, \dots, \vx_{k-1}]$ serves as the condition
for $\vx_k$. A naive way to train this is random-prefix causal fine-tuning: 
sample a split index $i$ per example, freeze chunks $<i$ at their clean values, and denoise chunk $i$
under a block-causal mask at one global noise level. Its drawback is that each
iteration supervises a single chunk.


\paragraph{Doubled token axis.}
We instead duplicate the token axis of each modality. With $\vx^{v}$, $\vx^{a}$
the clean video and audio latents, we form $\tilde{\vx}^{v} = [\,\vx^{v} \,;\,
\vx^{v}_{\sigma}\,]$ and $\tilde{\vx}^{a} = [\,\vx^{a} \,;\,
\vx^{a}_{\sigma}\,]$, where $\vx^{m}_{\sigma}$ is the noisy latent at level
$\sigma$. The clean and noisy latents are concatenated along the time dimension and share
identical rotary positions. Each noisy chunk $i$ draws its
own step index $s_i \sim \mathcal{U}\{0,\dots,999\}$ and reads the noise levels
$\sigma^v_i, \sigma^a_i$ from the video and audio schedules at that index, so
one forward pass sees every chunk at a different noise level. The two modalities share the
index $s_i$, keeping video and audio at the same stage of denoising within a chunk.

\paragraph{Mask and loss.}
A block-causal mask is applied to the attention matrix $A$ of the transformer.
A noisy chunk $i$ may attend its own noisy tokens and its clean history $< i$,
never the clean copy of chunk $i$.
Let $b(\cdot)$ be a row's chunk index and $c(\cdot)$ the indicator that it belongs
to the clean half. For query $q$ and key $k$, attention is permitted iff
\begin{equation}
\underbrace{c_q \wedge c_k \wedge (b_k \le b_q)}_{\text{clean context}}
\;\vee\;
\underbrace{\neg c_q \wedge \neg c_k \wedge (b_k = b_q)}_{\text{intra-chunk}}
\;\vee\;
\underbrace{\neg c_q \wedge c_k \wedge (b_k < b_q)}_{\text{clean history}},
\label{eq:mask}
\end{equation}
with text rows bidirectionally visible and conditioning rows at chunk index $-1$
so they precede all content. The second clause is what lets the two
modalities share knowledge inside a chunk. The third clause is the constraint
that matters: of the clean tokens, a noisy chunk reads only \emph{strictly
earlier} ground truth, never the clean copy of its own chunk, which sits at the
same rotary positions and would leak the answer. An attention mask example is
given in Figure~\ref{fig:mask}, Appendix~\ref{app:mask}. The loss
applies on the noisy half only, over chunks $i$ and modalities $m \in \{v,a\}$ with
the schedule's per-timestep weight $w_i$ and no extra coefficient between
modalities:
\begin{equation}
\gL = \sum_{m \in \{v,a\}} \sum_{i} w_i \big\| f_\theta(\tilde{\vx})^{m}_{i} - (\vepsilon^m_i - \vx^m_i) \big\|^2 .
\label{eq:loss}
\end{equation}
Eq.~\ref{eq:loss} is a plain flow-matching regression, which is what makes it
safe on the frozen undistilled weights and unsafe on the distilled ones.

\subsection{Why the Two Adapters Compose: An Orthogonality We Predict}
\label{sec:ortho}

The two adapters are \emph{added} at inference, so the sum is safe exactly if
their updates do not interfere; we argue they are near-orthogonal in what they
\emph{write}. Two structural reasons make this expected rather than fortunate,
and \S\ref{sec:geom} verifies it.

First, the two edit \emph{different functional axes}. $\Delta_d$ reparameterizes
the denoising map along the noise axis, that is, the network's output as a
function of the timestep $\sigma$. $\Delta_c$ instead edits \emph{context
routing}---how a token aggregates clean history through the cache under a
block-causal mask, a function of position and attention structure invariant to
the noise level. The two thus respond to different input factors: ``how much
noise'' versus ``which tokens are visible''.

Second, both updates are tiny perturbations of the frozen weights---$0.15\%$
of the projection norms for $\Delta_d$ and $0.28\%$ for $\Delta_c$
(\S\ref{sec:geom})---so the composed model stays in the small neighborhood of
$W$ where behavior is approximately linear, the regime where
\citet{ortizjimenez2023tangent} find independently trained task vectors
near-orthogonal by default. The two updates meet only in the attention
projections, the module type \citet{jin2025attention} find most disentangled in
task arithmetic, which is exactly where \S\ref{sec:geom} measures them.

This is a prediction: to second order the entire cost of adding the two
updates is their Hessian cross-term, which vanishes for $H$-orthogonal updates
(Appendix~\ref{app:secondorder}). \S\ref{sec:geom} confirms it, finding that
the adapters \emph{read} overlapping inputs yet \emph{write} into
near-orthogonal directions---the orthogonality is intrinsic, not \emph{imposed}
as in LoRA-merging recipes \citep{zhang2025osrm}.

\subsection{Weight Fusion and Autoregressive Inference}
\label{sec:cache}

\paragraph{Parallel composition by weight fusion.}
The two adapters are combined by weight arithmetic
\citep{ilharco2023taskarithmetic}: $\Delta_d$ is fused into the backbone offline
($W \leftarrow W + BA$), and $\Delta_c$ is loaded on top at inference, so
$\Delta_d$ adds no runtime overhead. One might expect failure---$\Delta_d$'s
trajectory was learned under full-sequence bidirectional attention, yet it now
runs under a block-causal mask with history entering through a cache---but
\S\ref{sec:causalcost} shows that four-step sampling still holds there. The streaming machinery below is
shared by all causal configurations.

\paragraph{Position-anchored streaming.}
Streaming generates one chunk at a time inside a sliding context window of at
most $N$ chunks. Already-produced chunks stay frozen as context and, together
with the encoded first-frame anchor, form the KV cache. Only
the latest chunk goes through the DiT; it is emitted after denoising, and one
extra forward writes its clean version into the cache. Every window starts at
frame zero, so a chunk keeps the same absolute rotary positions wherever it
appears and a cached key stays valid. Once the window holds $N$ chunks, the
oldest chunk leaves while the first-frame anchor stays, so the per-chunk cost
does not grow with session length (\S\ref{sec:cost}). The text prompt is the
one part that is never cached. Teacher forcing lets it attend every clean
chunk during training, while at inference each window re-encodes it and recomputes its keys alongside
the current chunk, a negligible cost at ${\sim}500$ tokens. The history it
reads back was encoded under the earlier, shorter contexts of the windows
that produced it; we accept this mismatch with training and call it the
\emph{stale cache}.

\section{Experiments}
\label{sec:experiments}

\subsection{Experimental Setup}
\label{sec:setup}

\noindent\textbf{Implementation details.} We initialize our model from the
official MiniMax-H3 FL2VA backbone and train it on avatar clips lasting
$5$--$10$~s in two resolution buckets, $480\!\times\!832$ and
$640\!\times\!640$ (Appendix~\ref{app:data}). The backbone stays frozen and both adapters are rank-$128$
LoRAs~\citep{hu2022lora}. We do not train
$\Delta_d$: it is the public lightx2v Turbo LoRA~\citep{lightx2v2026turbo},
which fixes the sampler at $4$ steps. We train $\Delta_c$ alone by
clean-context \cctf{} on chunks of $F\!=\!34$ frames, with
AdamW at a constant learning rate of $10^{-4}$ on sixteen GPUs for two
epochs. At inference, our method uses four denoising steps for each
audio--video chunk with a context window of $N=7$ chunks.

\noindent\textbf{Evaluation protocol.} We construct a held-out benchmark
spanning both resolution buckets, two languages and a range of
utterance lengths (Appendix~\ref{app:data}). No test clip is seen during training. Comparative
evaluations on this set are paired to the teacher's own generation of the same
clip, so reference and generation are comparable frame by frame. We further evaluate on three
public talking-head sets --- AVSpeech, HDTF and VidChatBench ($100$ clips
each) --- scored against the real source rather than the teacher, and on
$30$\,s rollouts, with history capped by
the sliding window. The held-out numbers measure fidelity to the teacher on
unseen prompts and
are not comparable with the public-set results. All systems share the
same first-frame-conditioned audio--video setting.

\noindent\textbf{Baselines.} The baselines all build on the same backbone $W$
and differ only in which adaptation they carry and in what order it was
acquired. \textsc{Teacher} ($50$-step, bidirectional) is a strong offline
quality reference but cannot emit block-causally. \textsc{Whole-seg}
($W\!+\!\Delta_d$, $4$-step) isolates distillation in its native bidirectional
mode. The chained orders interleave the two adaptations instead of separating
them: \textsc{Chained-DA} causalizes the distilled weights with a plain
regression loss, while \textsc{Chained-AD} follows the causal-forcing order
\citep{zhu2026causalforcing,zhao2026causalforcingpp}, for which we report the
anchored checkpoint. \textsc{Decoupled-50} serves our $\Delta_c$ on the
undistilled weights at $50$ steps and is a reference point, not a step
ablation. Our method is \textsc{Decoupled}
($W\!+\!\Delta_d\!+\!\Delta_c$, $4$-step, chunkwise). The columns of
Table~\ref{tab:main} form a ladder:
\textsc{Teacher}$\to$\textsc{Whole-seg} is distillation,
\textsc{Whole-seg}$\to$\textsc{Decoupled} is causalization with steps, shift,
backbone and seed held fixed, and \textsc{Decoupled}$\to$\textsc{Chained-DA}
is chaining.

\noindent\textbf{Metrics.} Quality metrics are scored per clip and paired
against the matching reference.
Audio--video synchronization is measured by Sync-C ($\uparrow$) and Sync-D
($\downarrow$) for lip--speech alignment~\citep{chung2016out}, and by AV-IB
($\uparrow$)~\citep{girdhar2023imagebind} and DeSync
($\downarrow$)~\citep{iashin2024synchformer} for audio--video semantics.
Visual quality uses IQA ($\uparrow$, CLIP-IQA style~\citep{wang2023clipiqa} on
ViT-L/14~\citep{radford2021clip}) and ASE ($\uparrow$)~\citep{schuhmann2022laion}
for appearance, CLIP ($\uparrow$) for prompt alignment, and CSIM ($\uparrow$),
the cosine similarity of detected-face embeddings against the clip's first
frame, for identity. Intelligibility is scored from
Qwen3-ASR-1.7B~\citep{qwen3asr2025} transcripts: WER ($\downarrow$) on English
and CER ($\downarrow$) on Chinese.
Long-horizon stability is reported as drift, the last window over the first,
for IQA and CSIM. Table~\ref{tab:main} reports these per-clip metrics;
Sync-D, DeSync and CSIM-r (identity against the reference portrait, not the
clip's own first frame) enter the public-set comparison
(Appendix~\ref{app:public}).
The set-level FID~\citep{heusel2017fid}, FVD~\citep{unterthiner2018fvd} and
FAD~\citep{kilgour2019fad} are computed on the $640\!\times\!640$ bucket
against unseen teacher output and quoted in the text beside their floor, so
quality is the gap rather than an absolute value. Throughput, in frames per
second (FPS) of end-to-end wall-clock time, is in Appendix~\ref{app:cost}.
Implementation controls are in Appendix~\ref{app:metrics}.

\begin{table}[t]
\centering
\caption{Results on the held-out benchmark (\S\ref{sec:setup}). Drift rows are
last window over first; CSIM-1st is identity against the clip's first frame. Speech is transcribed with
Qwen3-ASR-1.7B: WER for the English clips, CER for the Chinese ones. Among the
four autoregressive systems, best results in bold, second-best underlined.}
\label{tab:main}
\footnotesize
\setlength{\tabcolsep}{4.0pt}
\begin{tabular}{lcccccc}
\toprule
 & \textsc{Teacher} & \textsc{Whole-seg} & \textsc{Decoupled} & \textsc{Decoupled} & \textsc{Chained-DA} & \textsc{Chained-AD} \\
 & $50$ step & $4$ step & $4$ step (ours) & $50$ step & $4$ step & $4$ step \\
\midrule
Sync-C $\uparrow$ & $8.412$ & $8.403$ & \underline{$7.443$} & $7.197$ & $6.389$ & $\mathbf{8.021}$ \\
AV-IB $\uparrow$ & $0.414$ & $0.414$ & $0.403$ & $\mathbf{0.423}$ & $0.363$ & \underline{$0.419$} \\
\midrule
IQA $\uparrow$ & $0.908$ & $0.915$ & $\mathbf{0.914}$ & \underline{$0.894$} & $0.854$ & $0.877$ \\
ASE $\uparrow$ & $5.048$ & $5.110$ & $\mathbf{5.106}$ & \underline{$5.023$} & $5.019$ & $5.008$ \\
CLIP $\uparrow$ & $0.212$ & $0.212$ & $\mathbf{0.212}$ & $0.210$ & $\mathbf{0.212}$ & $\mathbf{0.212}$ \\
CSIM-1st $\uparrow$ & $0.886$ & $0.870$ & \underline{$0.864$} & $0.853$ & $\mathbf{0.871}$ & $0.835$ \\
IQA drift ($\to 1$) & $1.005$ & $1.008$ & $1.030$ & $1.026$ & $\mathbf{1.002}$ & \underline{$0.990$} \\
CSIM drift ($\to 1$) & $0.981$ & $0.981$ & $\mathbf{0.950}$ & \underline{$0.948$} & $0.945$ & $0.916$ \\
\midrule
WER (En) $\downarrow$ & $0.041$ & $0.047$ & $0.127$ & $\mathbf{0.110}$ & \underline{$0.126$} & $0.136$ \\
CER (Zh) $\downarrow$ & $0.014$ & $0.008$ & $\mathbf{0.051}$ & \underline{$0.055$} & $0.061$ & $0.061$ \\
\bottomrule
\end{tabular}
\end{table}

\subsection{Main Results}
\label{sec:distfree}
\label{sec:causalcost}
\label{sec:chained}

\noindent\textbf{Audio--visual quality on the held-out benchmark.}
Table~\ref{tab:main} collects the per-clip metrics, which we analyze from two
perspectives. First, distillation alone: the bidirectional four-step
\textsc{Whole-seg} stays within measurement noise of the $50$-step
\textsc{Teacher} on nearly every metric (Sync-C $8.403$ against $8.412$,
identical AV-IB, IQA and ASE even slightly above the teacher), so ``four
steps is too few'' explains none of the gaps that follow. Second, we compare
the four-step autoregressive systems on appearance, synchronization and
speech.
\label{sec:speechsec}%
\emph{Appearance.} \textsc{Decoupled} keeps the image quality of
\textsc{Whole-seg} (IQA $0.914$ against $0.915$), ahead of both chains
($0.854$ and $0.877$), and holds identity best over the clip (CSIM drift
$0.950$); \textsc{Chained-DA} is slightly ahead on first-frame identity.
\emph{Synchronization.} This is the axis a chain wins: \textsc{Chained-AD},
which distills on the causal model itself, has the best lip sync (Sync-C
$8.021$ against our $7.443$) and pays for it in appearance and identity;
\textsc{Decoupled} stays well ahead of \textsc{Chained-DA} ($6.389$).
\emph{Speech.} Causalization costs most here. Every four-step causal system
makes close to three times the word errors of \textsc{Whole-seg} (WER
$0.126$--$0.136$ against $0.047$): chunkwise attention lets an audio token
attend only the audio already generated, so the model must infer how far into
the script it is from that alone. \textsc{Decoupled} matches the best chain on English (WER $0.127$
against $0.126$ and $0.136$), has the lowest Chinese error rate (CER $0.051$
against $0.061$ for both chains), and keeps an audio distribution closer to the
teacher than \textsc{Chained-DA}'s (FAD $3.19$ against $5.02$).

\textsc{Chained-DA} falls behind on each axis, and its FID rises above the
teacher's floor ($70.31$ against $67.10$) while \textsc{Decoupled} stays below
it ($66.15$), the un-distillation \S\ref{sec:routes} predicts. Composition is
thus the strongest four-step streaming configuration in the table. At
inference, the stream sustains ${\approx}26$\,fps on six GPUs
(Appendix~\ref{app:cost}).

\noindent\textbf{Further validation on public benchmarks.}
\label{sec:external}
The held-out conclusions carry over to three public benchmarks ($100$ clips
each), scored against the real source rather than the teacher
(Table~\ref{tab:external}, Appendix~\ref{app:public}).
\textsc{Decoupled} places first or second among the four autoregressive
systems on nearly every scored column: IQA $0.866$/$0.866$/$0.892$ on
AVSpeech, VidChatBench and HDTF, ahead of every autoregressive baseline and
of the $50$-step teacher on all three; the best autoregressive AV-IB on every
set; and the lowest autoregressive FID and FVD on AVSpeech and VidChatBench.

\begin{figure}[t]
\centering
\includegraphics[width=0.95\linewidth]{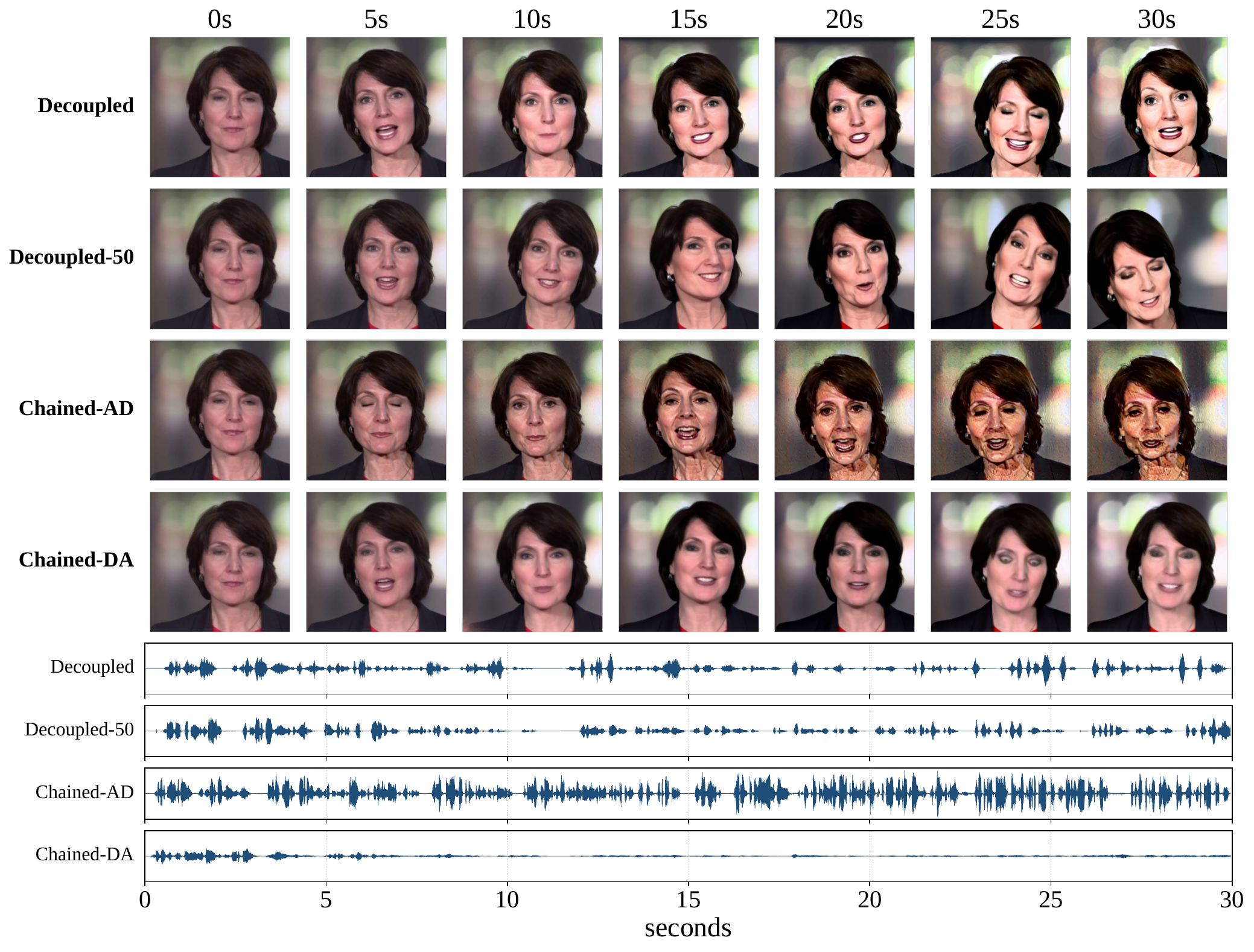}
\caption{One clip streamed to $30$\,s. Rows are systems and columns are wall-clock seconds; below them the
four generated soundtracks share one time axis. \textsc{Chained-DA} stops
articulating well before the end while the composed routes speak throughout;
Table~\ref{tab:longrun-windows} reports the same comparison over all forty clips.}
\label{fig:longrun}
\end{figure}

\noindent\textbf{Long-horizon analysis.}
\label{sec:accumulate}
What remains to be settled is whether composition survives
\emph{extrapolation}. We stream to $30$\,s, well past the $5$--$10$\,s clips
the causal adapter was trained on, with no long-horizon training of any kind. We evaluate the
resulting long-horizon behavior two ways --- by \emph{endpoint change},
scoring the closing window against the opening one, and by the full
\emph{temporal trajectory}, tracking first-frame similarity densely along the
clip. Both use the same $40$ clips, whose public-set transcripts we extend
until each carries more script than $30$\,s of speech can deliver, so that every
system is still mid-sentence when the video ends and none can be rewarded for
simply stopping, and we generate the full $30$\,s with the history capped by
the sliding window (one clip is shown in Figure~\ref{fig:longrun}).

\begin{figure}[t]
\centering
\includegraphics[width=\linewidth]{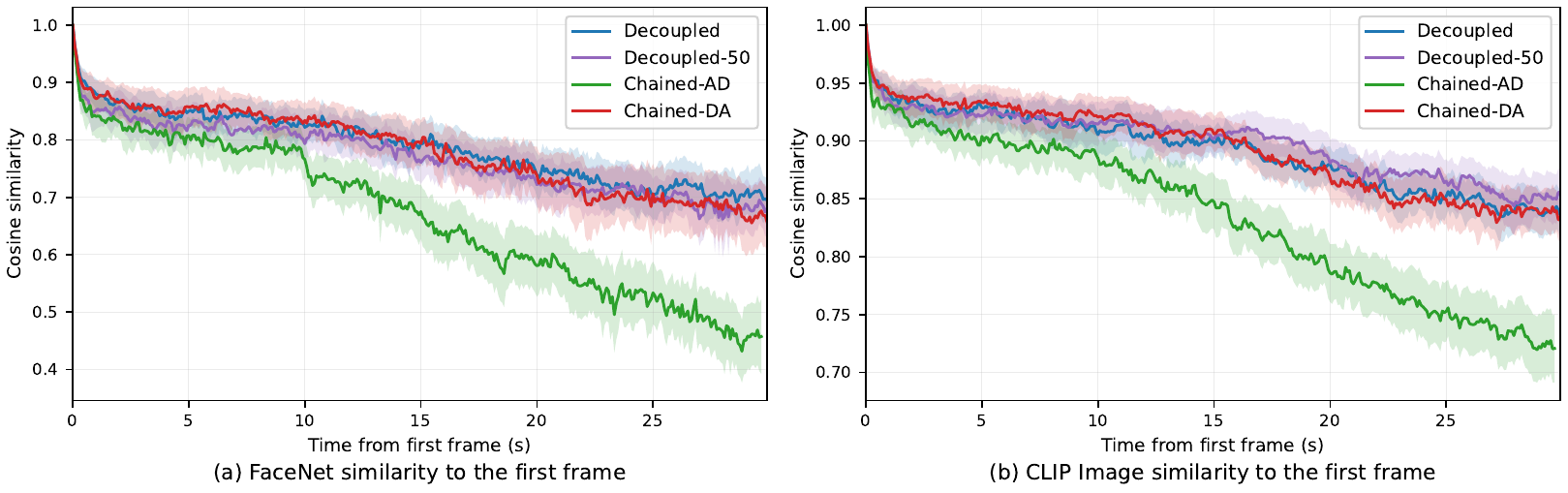}
\caption{Long-horizon consistency as a function of generation time.
(a)~FaceNet similarity and (b)~CLIP Image similarity to the first frame of
the same video. Curves are means over $40$ clips with $95\%$ confidence
intervals, and terminate at each method's actual duration.}
\label{fig:consistency}
\end{figure}

\label{sec:longrun}
Table~\ref{tab:longrun-windows} reports each metric over the final $5$-second
window together with its relative change from the initial $5$ seconds.
\textsc{Decoupled} closes best
on identity and image quality --- $0.710$ and $0.871$, the latter
$2.6\%$ \emph{above} its own opening --- while \textsc{Chained-AD} arrives at
$0.484$ and $0.526$, having lost over a third of each, and
\textsc{Chained-DA} keeps its face ($0.683$) but nearly halves its lip sync
($3.75$, $-43.7\%$). The same holds for audio. Measured on raw samples, the closing
amplitude drops only $10.6\%$ from its opening for \textsc{Decoupled} against
$37.9\%$ for \textsc{Chained-DA}, which has largely fallen silent by the end. Two endpoints go against us: lip sync, where
\textsc{Chained-AD} closes highest ($7.82$ against $5.69$), and speaker
identity, which falls furthest for \textsc{Decoupled} ($0.521$, $-41.0\%$) and
which \S\ref{sec:limitations} reports as a limitation.
The first-frame trajectories (Figure~\ref{fig:consistency}) show the same pattern: similarity to the first
frame declines gradually under composition, with no late acceleration, while
\textsc{Chained-AD} falls faster after $10$\,s and \textsc{Chained-DA} stays
close to us in first-frame similarity, its cost showing in the audio and in
image quality instead.

\begin{table}[t]
\centering
\caption{Long-horizon preservation over a $40$-clip, $30$\,s rollout. Each cell is the score on the final $5$\,s and, in parentheses, its relative change from the opening $5$\,s. Both windows are the same length, so Sync-C is comparable between them but not with a full-clip Sync-C. Bold marks the best closing window; for RMS, which is an absolute level, it marks the smallest relative change instead.}
\label{tab:longrun-windows}
\footnotesize
\setlength{\tabcolsep}{4pt}
\begin{tabular}{@{}lcccccc@{}}
\toprule
Method & Sync-C $\uparrow$ & Identity $\uparrow$ & CLIP $\uparrow$ & IQA $\uparrow$ & Voice $\uparrow$ & RMS \\
\midrule
\textsc{Decoupled} & 5.69 ($-$29.6\%) & \textbf{0.710 ($-$18.3\%)} & 0.844 ($-$9.7\%) & \textbf{0.871 ($+$2.6\%)} & 0.521 ($-$41.0\%) & \textbf{0.086 ($-$10.6\%)} \\
\textsc{Decoupled-50} & 6.23 ($-$19.1\%) & 0.687 ($-$18.9\%) & \textbf{0.856 ($-$8.2\%)} & 0.848 ($+$3.0\%) & \textbf{0.682 ($-$22.3\%)} & 0.109 ($-$16.6\%) \\
\textsc{Chained-AD} & \textbf{7.82 ($-$12.8\%)} & 0.484 ($-$41.8\%) & 0.737 ($-$19.7\%) & 0.526 ($-$35.4\%) & 0.608 ($-$31.9\%) & 0.185 ($+$49.0\%) \\
\textsc{Chained-DA} & 3.75 ($-$43.7\%) & 0.683 ($-$21.4\%) & 0.841 ($-$10.5\%) & 0.656 ($-$16.2\%) & 0.603 ($-$32.3\%) & 0.047 ($-$37.9\%) \\
\bottomrule
\end{tabular}
\end{table}

\subsection{The Predicted Orthogonality, Verified}
\label{sec:geom}

\begin{table}[t]
\caption{Geometry of each causal update against each distillation update, over
all shared (block, projection) pairs, mean $\pm$ s.d. \textsc{Decoupled}
and \textsc{Chained-DA} measure $\Delta_d$ against $\Delta_c$ and $\Delta_c'$;
\textsc{Chained-AD} measures $\Delta_c$ against its two distillation increments
taken together ($\Delta_{cd} \oplus \Delta_{dmd}$). Interference and projection
are defined in Appendix~\ref{app:lorageom}. Projection is a signed alignment,
zero meaning none; both interference forms vanish for orthogonal subspaces, so
for all three quantities the null column is the value to read against rather
than zero.}
\label{tab:geom}
\footnotesize
\centering
\setlength{\tabcolsep}{5.0pt}
\begin{tabular}{lcccc}
\toprule
& \textsc{Decoupled} & \textsc{Chained-DA} & \textsc{Chained-AD} & null \\
\midrule
Projection              & $+0.0006 \pm 0.0055$ & $-0.0022 \pm 0.0084$ & $-0.0008 \pm 0.0349$ & $+0.0000 \pm 0.0008$ \\
Write interference      & $0.0147 \pm 0.0126$ & $0.0169 \pm 0.0159$ & $0.0451 \pm 0.0308$ & $0.0117 \pm 0.0042$ \\
Read interference       & $0.1356 \pm 0.1058$ & $0.1061 \pm 0.1027$ & $0.1844 \pm 0.1094$ & $0.0237 \pm 0.0180$ \\
\bottomrule
\end{tabular}
\end{table}

\S\ref{sec:ortho} predicted that the two adapters write into near-orthogonal
directions. We test it on the trained weights alone, with no generation
involved: Table~\ref{tab:geom} measures each causal update against each
distillation update over the attention projections both adapters edit, against
a null that pairs each block's update with another depth's
(Appendix~\ref{app:lorageom}); comparisons are paired signed-rank tests over
those pairs. Both edits are tiny (\S\ref{sec:ortho}), so the composed point is
a small perturbation of $W$. On that scale the signed alignment between the two updates
is not separated from the null by the paired test ($+0.0006$ against
$+0.0000$, $p{=}0.26$); write interference exceeds the null by $26\%$ ($0.0147$
against $0.0117$; measurable at $p{=}5{\times}10^{-4}$ but small) while read
interference exceeds it by $472\%$ ($5.7\times$, $p{<}10^{-30}$) --- a factor
of eighteen between the two excesses. The adapters
therefore \emph{read} overlapping inputs yet \emph{write} into near-orthogonal
directions, as predicted: adding them does not overwrite either behavior, so
the two tasks can be trained independently and composed by a sum. The $52\%$ of $\Delta_d$ in feed-forward projections
that $\Delta_c$ never touches is orthogonal by construction and excluded from the
table, leaving only the potentially conflicting projections and making the test
conservative.

The same numbers show that neither chain fails for geometric reasons.
\textsc{Chained-DA}'s $\Delta_c'$ is geometrically almost indistinguishable
from ours --- both projections lie within $0.003$ of zero alignment ($-0.0022$
against $+0.0006$), a gap the test resolves but one far too small to carry a
behavioral difference --- so its fade cannot be read off the parameters. The
reverse order does drift: \textsc{Chained-AD}'s distillation increment, trained
on top of $\Delta_c$, has write interference $0.045$, three times ours
($p{<}10^{-25}$), plausibly because a critic trained on the causal model's own rollouts
corrects in directions correlated with what $\Delta_c$ wrote --- still far too
small to break a sum. Orthogonality is what the parallel route gets for free
from independent training, not a precondition any route needs; what separates
the routes is the objective.

\label{sec:flexible}
Write-orthogonality also makes $\Delta_c$ reusable without retraining: it writes
into directions $\Delta_d$ leaves empty, so it binds to neither the distillation
nor the step budget. Added to $\Delta_d$ it is the four-step
\textsc{Decoupled}; served alone on the undistilled weights at $50$ steps it
is \textsc{Decoupled-50} (Sync-C $7.197$, IQA $0.894$) --- one file on both
the distilled and the undistilled weights. \textsc{Chained-DA} retrains its
causal adapter so that training and serving weights match; ours needs no such
retraining.

\section{Limitations}
\label{sec:limitations}

\paragraph{Talking-head data only.} Every corpus we train or evaluate on ---
the adaptation set, the held-out split and the public benchmarks --- is
talking-head (digital-human) footage. The costs we attribute to causalization
--- lip--speech coupling, the
audio distribution --- and the metrics that measure them are therefore specific
to this domain.

\paragraph{Speaker identity drifts over long sessions.} On the long rollouts of
\S\ref{sec:longrun} the composed route holds \emph{face} identity as well as
any chain, but \emph{speaker} identity does not: voice similarity to the clip's
opening falls furthest for \textsc{Decoupled} (Figure~\ref{fig:similarity}).
$\Delta_d$ learned the audio field under whole-sequence attention; the composed
model must keep timbre from cached history alone, and our causal adapter never
sees its own rollouts. Closing that gap with a self-rollout objective is
natural future work.

\paragraph{A single backbone.} All experiments are on MiniMax-H3 (FL2VA). We
do not evaluate on other joint audio--video backbones (e.g.\ LTX-2 or the Wan
family), so whether the parallel composition transfers remains untested. We
also do not compare against OmniForcing \citep{su2026omniforcing}, the closest
published joint audio--video streaming method: it is trained on LTX-2, and a
cross-backbone table would confound the generator with the recipe. A
same-backbone comparison would require porting their chained pipeline onto
H3, or ours onto LTX-2.

\section{Conclusion}

In this paper, we showed that block-causal attention and few-step sampling
need not be learned in series. A causal adapter trained by clean-context
\cctf{} against the frozen backbone, added at inference to an off-the-shelf
few-step adapter, yields a few-step streaming audio--video model with no joint
training and no orthogonality regularizer. The two updates write into
near-orthogonal directions, as predicted from the functional axes they edit
and as measured on the trained weights, which is why the sum is safe.
Empirically, the composed model tracks the bidirectional teacher on
appearance and holds quality over rollouts longer than the adaptation clips.
The same causal adapter can be served on both the undistilled and the distilled
weights. A
natural next step is to test whether the same parallel composition holds on
other joint audio--video generators.\label{endconcl}

\bibliography{iclr2027_conference}
\bibliographystyle{iclr2027_conference}

\appendix
\section{Training and Held-Out Data}
\label{app:data}
Our training and test data come from a corpus of around $4000$ clips of $5$--$10$\,s.
The clips are generated by MiniMax-H3 \citep{minimax2026h3} with its native $50$ steps,
using text prompts in talk-show and talking-head styles. The generated videos
are center-cropped and resized to fit the bucket sizes: $480{\times}832$ and $640{\times}640$.
$211$ clips are sampled from this corpus as the held-out test set, and the remaining clips are used for training.
Both training and test data are balanced across two languages: Chinese and English.

For every held-out clip, frame counts follow the $17n{+}5$ grid, so
generation and reference compare frame by frame. The reference is the
backbone's own $50$-step generation of the same clip, which conditions on both
keyframes, whereas each streamed system sees only the first --- a strict upper
bound.

\section{Backbone, Chunk Geometry and the Stale Cache}
\label{app:geometry}

\paragraph{Backbone.}
We adapt MiniMax-H3 \citep{minimax2026h3} in its first-frame-conditioned
audio--video (FL2VA) configuration. Video is encoded by a causal VAE mapping
$17$ frames to $5$ latent tokens, so a legal clip length is $17n+5$ frames;
audio is a $40$\,Hz stereo latent. Text, a first-frame condition and both
modalities form one packed sequence with three-dimensional rotary
positions \citep{su2021roformer} and share the same blocks. The backbone is
frozen throughout.

\paragraph{Chunk geometry.}
The video VAE requires the chunk size $F$ to be a multiple of $17$ frames;
audio runs at $40$\,Hz against $24$\,fps, so a boundary is exact only when
$3 \mid F$. The conservative choice is $F=\mathrm{lcm}(17,3)=51$; we use
$F=34$. The audio index is taken from the absolute frame number rather than
accumulated per chunk, so the rounding error stays within a third of a latent
and does not grow.

\paragraph{Why we accept a stale cache.} Text rows are bidirectionally visible
(Eq.~\ref{eq:mask}), so training computes the text and every clean chunk in
one forward pass: the text attends the whole clip, and each history key is
produced in that full-clip context. Streaming cannot reproduce this. Each
window re-encodes the text alongside the current chunk, which is cheap at
${\sim}500$ tokens, but the history it reads from the cache was written by
earlier, shorter windows, under texts that had seen less of the clip. A cached
key is therefore stale in two ways: the text that helped encode it has moved
on, and the context it aggregated was a window rather than the clip. The
exact alternative is re-encoding the whole history at every window, which is
the whole-segment cost the cache exists to avoid; we re-encode the cheap part
and accept the staleness.

\section{Mask Verification}
\label{app:mask}
\begin{figure}[t]
\centering
\includegraphics[width=0.9\linewidth]{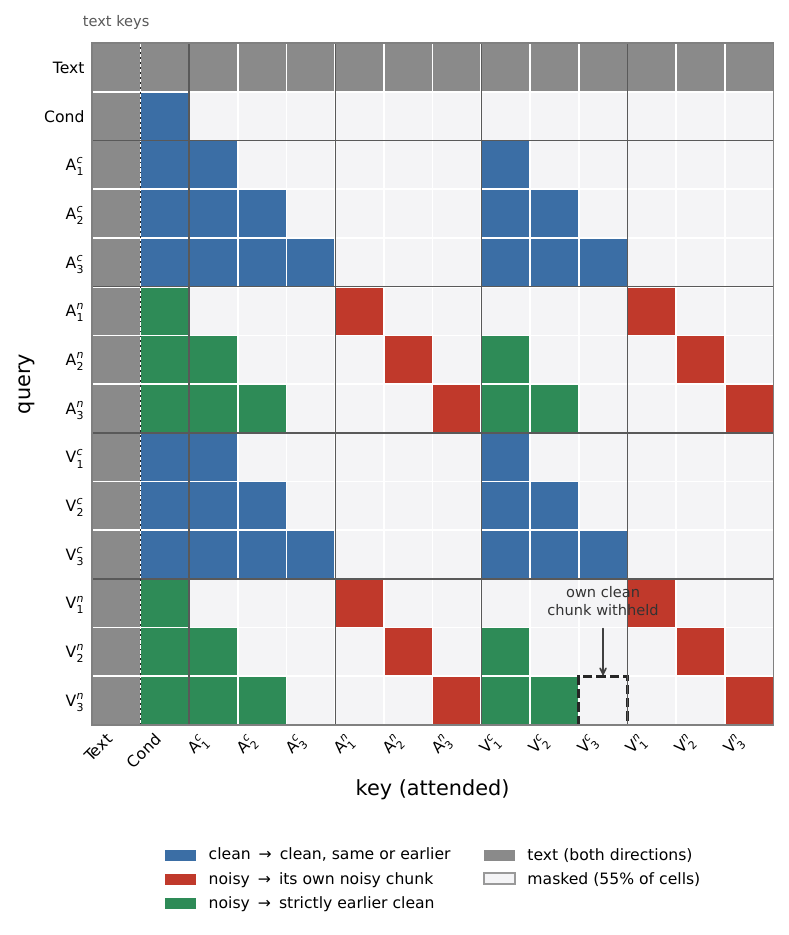}
\caption{Attention pattern of Eq.~\ref{eq:mask} on a three-chunk packed
sequence. Clean tokens attend same-or-earlier clean tokens; a noisy chunk
attends its own noisy tokens and strictly earlier clean history, never its
clean copy. Conditioning is chunk $-1$: visible as a key to every content
row, but a conditioning query cannot read later clean tokens. Audio is
drawn chunkwise, as video is.}
\label{fig:mask}
\end{figure}

A noisy chunk attending its own clean copy lowers the loss without being
noticed, so we check the layout in code: condition rows are clean and visible
to every chunk, and no noisy row attends a clean row of its own or a later
chunk. Figure~\ref{fig:mask} shows the permitted pattern.

\section{Metric Implementation Controls}
\label{app:metrics}

Metric names are defined in \S\ref{sec:setup}. We record here only the
choices that change the numbers or how they should be read.

IQA uses CLIP-IQA-style antonym prompting on ViT-L/14 rather than the
published RN50, so the scores are not comparable across papers. Sync-C and
Sync-D follow SyncNet on a $25$\,fps / $100$\,Hz MFCC grid, with BGR input in
$0$--$255$ and MFCCs on int16 amplitudes. DeSync's $21$ classes are $200$\,ms
apart, so it is the right metric for cross-paper alignment and the wrong one
for offsets of a few tens of milliseconds. AV-IB distinguishes speech from
non-speech but barely whose speech it is; on this corpus it is a sanity check.

CSIM in Table~\ref{tab:main} and Figure~\ref{fig:consistency} uses FaceNet
(VGGFace2) under MTCNN, taking the highest-confidence box.
Figure~\ref{fig:similarity} uses ArcFace and should not be compared in
absolute cosine. Speech is transcribed with Qwen3-ASR-1.7B, language forced
from the prompt: WER on English, CER on Chinese, after folding case and
stripping punctuation, so an early finish is counted as deletions.

\section{Why Addition Works: A Second-Order View}
\label{app:secondorder}
Let $f$ be the training loss, $W$ the frozen weights, and $\Delta_c$,
$\Delta_d$ the two LoRA updates. A second-order expansion of the loss of the
composed model around $W$ reads
\[
f(W+\Delta_c+\Delta_d) \;=\; f(W+\Delta_c) + f(W+\Delta_d) - f(W)
\;+\; \Delta_c^{\!\top} H\, \Delta_d \;+\; O(\|\Delta\|^3),
\]
with $H$ the Hessian at $W$. The cross-term
$\Delta_c^{\!\top} H\, \Delta_d$ collects every interaction between the two
updates: in this approximation, \emph{the entire cost of combining them is
the cross-term}, and it vanishes when the updates are orthogonal in the
metric of $H$.

The measurements of \S\ref{sec:geom} estimate this cross-term in the
Frobenius metric. The signed alignment between the two updates is not
separated from the null, and write interference exceeds it only modestly, so
the written directions are no more aligned than two random low-rank subspaces
of the same shape. That bounds the cross-term under an isotropic
approximation to $H$. Frobenius orthogonality is necessary, not sufficient,
for $H$-orthogonality, so we read it as an estimate, not a proof. Read
interference is a different quantity: it concerns overlapping inputs, not the
Hessian cross-term.

This is the structure weight averaging exploits from the other side. Model
soups \citep{wortsman2022modelsoups} average multiple fine-tunes of one base
and gain when those solutions share a basin in which the loss is locally
near-convex --- the linear mode connectivity observed for solutions of a
common initialization \citep{frankle2020linear}. Our updates differ in both
respects: they are trained for \emph{heterogeneous} objectives (bidirectional
few-step distillation against block-causal causalization), and only one of
them is trained at all --- the other is taken off the shelf. No
basin-sharing argument is available; what the sum needs is precisely that
the cross-term be small, which is the quantity \S\ref{sec:geom} measures.

\section{How the Adapter Geometry Is Computed}
\label{app:lorageom}
Both adapters are low-rank, so every quantity in Table~\ref{tab:geom} follows from
$r\times r$ matrices. For $X=B_xA_x$ and $Y=B_yA_y$ with $M=B_x^{\!\top}B_y$,
$G_x=A_xA_x^{\!\top}$ and $H_x=B_x^{\!\top}B_x$,
\begin{equation*}
\langle X,Y\rangle_F=\operatorname{tr}\!\big(M(A_yA_x^{\!\top})\big),\quad
\|X\|_F^2=\operatorname{tr}(H_xG_x),\quad
\|X^{\!\top}Y\|_F^2=\operatorname{tr}(M^{\!\top}G_xMG_y),
\end{equation*}
and $\|XY^{\!\top}\|_F^2=\operatorname{tr}(N^{\!\top}H_xNH_y)$ with
$N=A_xA_y^{\!\top}$. The last two vanish exactly when the column, respectively
row, spaces are orthogonal. Subspace overlap is $\|Q_x^{\!\top}Q_y\|_F^2/r$ for
orthonormal bases from a QR of $A^{\!\top}$; for two random $r$-dimensional
subspaces of $\mathbb{R}^{n}$ its expectation is $r/n$. The $p$-values quoted
in \S\ref{sec:geom} are two-sided Wilcoxon signed-rank tests over the paired
(block, projection) values, each route against the null computed on the same
pair.

One asymmetry between the two causal updates does not show up in
Table~\ref{tab:geom}: $\Delta_c'$ is $27\%$ smaller than $\Delta_c$ in Frobenius
norm overall and only $44\%$ its size averaged over the first quarter of the
blocks, against $94\%$ over the last quarter, so training on $W'$ spends its
budget late in the network rather than early.

Both adapters are compared on the attention projections they share. About half
of $\Delta_d$'s parameters sit in the feed-forward layers, which $\Delta_c$ does
not touch; those layers are orthogonal by construction, are excluded from
Table~\ref{tab:geom}, and are not part of the write-interference claim. Norms
are reported per (block, projection) as $\|\Delta\|_F/\|W\|_F$ and then averaged.

\section{Public Benchmarks}
\label{app:public}

Table~\ref{tab:external} is the full comparison summarized in
\S\ref{sec:external}: the systems of Table~\ref{tab:main} on three public
sets, scored against the real source.

\begin{table}[h]
\centering
\caption{The systems of Table~\ref{tab:main} on three public test sets ($100$
clips each), scored against the \emph{real source video} --- so \textsc{Teacher}
is a system under test, not the reference. Among the four autoregressive
systems, best results in bold, second-best underlined, per column and set; the
GT row is a reference, not a competitor. CSIM-r is face identity against the
reference portrait, not CSIM-1st. FAD is omitted on HDTF: clip-length
mismatch dominates there (whole-segment generation has a $4$\,s minimum against
${\sim}3.2$\,s sources); DeSync is undefined on HDTF (clips shorter than
Synchformer's $4.8$\,s context).}
\label{tab:external}
\footnotesize
\setlength{\tabcolsep}{3.0pt}
\begin{tabular}{lccccccccc}
\toprule
  & IQA $\uparrow$ & Sync-C $\uparrow$ & Sync-D $\downarrow$ & AV-IB $\uparrow$ & DeSync $\downarrow$ & CSIM-r $\uparrow$ & FID $\downarrow$ & FVD $\downarrow$ & FAD $\downarrow$ \\
\midrule
\multicolumn{10}{l}{\textbf{AVSpeech}~($n{=}100$)} \\
\emph{GT (source)} & 0.852 & 6.82 & 7.77 & 0.377 & 0.115 & 0.867 & --- & --- & --- \\
\cmidrule(l){2-10}
\textsc{Teacher} & 0.857 & 8.49 & 7.26 & 0.333 & 0.028 & 0.839 & 32.2 & 128.5 & 6.85 \\
\textsc{Whole-seg} & 0.835 & 7.82 & 7.62 & 0.328 & 0.040 & 0.864 & 33.7 & 119.2 & 5.96 \\
\textsc{Decoupled} & \textbf{0.866} & \underline{7.54} & \underline{7.85} & \textbf{0.334} & \textbf{0.007} & \textbf{0.791} & \textbf{35.2} & \textbf{129.0} & \underline{5.23} \\
\textsc{Decoupled-50} & \underline{0.836} & 7.13 & 7.99 & \underline{0.324} & 0.028 & 0.761 & 38.9 & 143.6 & 5.34 \\
\cmidrule(l){2-10}
\textsc{Chained-AD} & 0.804 & \textbf{8.39} & \textbf{7.63} & 0.305 & \underline{0.012} & 0.741 & \underline{38.8} & 150.0 & 7.04 \\
\textsc{Chained-DA} & 0.803 & 6.10 & 8.56 & 0.290 & 0.023 & \underline{0.790} & 40.2 & \underline{131.9} & \textbf{4.94} \\
\addlinespace
\multicolumn{10}{l}{\textbf{VidChatBench}~($n{=}100$)} \\
\emph{GT (source)} & 0.770 & 5.23 & 8.83 & 0.404 & 0.081 & 0.797 & --- & --- & --- \\
\cmidrule(l){2-10}
\textsc{Teacher} & 0.769 & 7.32 & 7.80 & 0.328 & 0.022 & 0.725 & 38.1 & 171.4 & 6.07 \\
\textsc{Whole-seg} & 0.712 & 6.69 & 8.29 & 0.319 & 0.022 & 0.752 & 40.5 & 178.9 & 8.20 \\
\textsc{Decoupled} & \textbf{0.866} & \underline{7.84} & \textbf{7.65} & \textbf{0.353} & \textbf{0.024} & \underline{0.774} & \textbf{39.6} & \textbf{184.0} & 9.05 \\
\textsc{Decoupled-50} & \underline{0.829} & 7.65 & \underline{7.69} & \underline{0.348} & 0.036 & 0.758 & \underline{41.3} & \underline{191.5} & \underline{6.66} \\
\cmidrule(l){2-10}
\textsc{Chained-AD} & 0.742 & \textbf{7.85} & 7.84 & 0.336 & \underline{0.031} & 0.698 & 42.4 & 203.0 & \textbf{6.19} \\
\textsc{Chained-DA} & 0.776 & 6.70 & 8.25 & 0.302 & 0.034 & \textbf{0.775} & 43.5 & 193.8 & 8.10 \\
\addlinespace
\multicolumn{10}{l}{\textbf{HDTF}~($n{=}100$)} \\
\emph{GT (source)} & 0.877 & 8.24 & 6.72 & 0.330 & --- & 0.937 & --- & --- & --- \\
\cmidrule(l){2-10}
\textsc{Teacher} & 0.874 & 8.94 & 6.66 & 0.297 & --- & 0.919 & 13.0 & 63.6 & --- \\
\textsc{Whole-seg} & 0.877 & 8.65 & 6.82 & 0.263 & --- & 0.933 & 11.3 & 48.3 & --- \\
\textsc{Decoupled} & \textbf{0.892} & \underline{9.04} & \underline{6.74} & \textbf{0.295} & --- & \underline{0.905} & \underline{14.4} & 61.4 & --- \\
\textsc{Decoupled-50} & \underline{0.880} & 8.69 & \textbf{6.72} & 0.270 & --- & 0.902 & 15.8 & \underline{59.2} & --- \\
\cmidrule(l){2-10}
\textsc{Chained-AD} & 0.853 & \textbf{9.28} & 6.76 & \underline{0.293} & --- & 0.869 & 22.3 & 98.9 & --- \\
\textsc{Chained-DA} & 0.876 & 8.58 & 6.86 & 0.253 & --- & \textbf{0.919} & \textbf{14.2} & \textbf{49.6} & --- \\
\addlinespace
\bottomrule
\end{tabular}
\end{table}

\section{Long-Horizon Sample}
\label{app:external}
\begin{figure}[H]
\centering
\includegraphics[width=\linewidth]{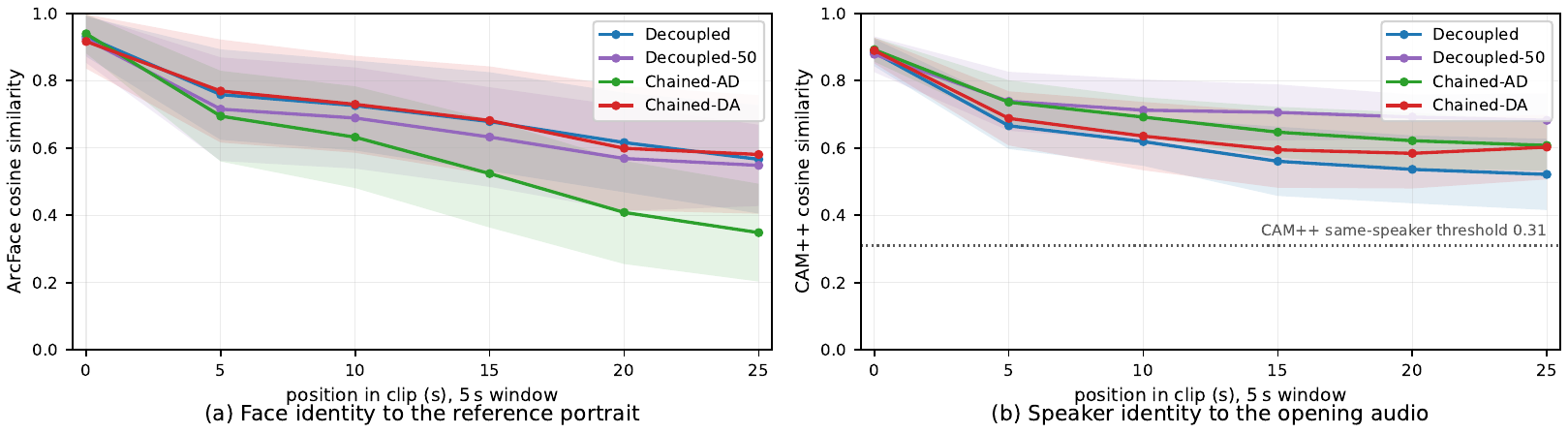}
\caption{Identity drift over the $30$\,s long-horizon runs of
\S\ref{sec:longrun} ($40$ clips, probes every $5$\,s).
(a)~Face similarity to the reference portrait, with ArcFace.
(b)~Speaker similarity to the clip's own opening segment, with CAM++. Both are the
complement of Figure~\ref{fig:consistency}, which measures against each clip's
own first frame under different encoders. Face identity is retained equally
well by \textsc{Decoupled},
\textsc{Decoupled-50} and \textsc{Chained-DA} (and worst by \textsc{Chained-AD});
speaker identity falls furthest for \textsc{Decoupled}, the observation
\S\ref{sec:limitations} reports as a limitation.}
\label{fig:similarity}
\end{figure}

The $30$\,s rollouts in \S\ref{sec:longrun} are summarized in the main text by
endpoint change. This appendix adds face and speaker similarity to a
fixed reference (Figure~\ref{fig:similarity}): face identity is retained
under composition, and speaker identity is the drift
\S\ref{sec:limitations} records.

\section{Inference Cost}
\label{app:costsec}
\label{app:cost}
\label{sec:cost}
The extra wall-clock of chunk-autoregressive inference is the key/value
prefill of \S\ref{sec:cache}, not the denoising steps: subtract it and the two
modes agree to a few percent. Time-to-first-frame is the first window,
$4.2$--$4.3$\,s at every length, against a whole-segment call that grows from
$11.2$\,s to $54.3$\,s. Within a clip both modes are quadratic in length;
across a session the sliding window keeps the per-chunk cost constant. On six
GPUs, without quantization, the stream sustains ${\approx}26$ frames
per second at $480{\times}832$, ahead of $24$\,fps playback.
\textsc{Decoupled} and \textsc{Chained-DA} share an adapter shape, so the same
timings apply to both.

\section{The Reverse Order: \textsc{Chained-AD}'s Training Stages}
\label{app:reverse}
\label{sec:reverse}
Stage~1 of \textsc{Chained-AD} is the same $\Delta_c$ as in the parallel route
(\S\ref{sec:cctf}); two further LoRA stages then distill the few-step property.
Neither of them applies a flow-matching regression to already-distilled
weights. We report the anchored checkpoint, served with the same streaming
inference as the other autoregressive systems.

Stage~2 is consistency distillation on the ground-truth cache: the student
matches a nearby point on the frozen autoregressive teacher's Euler
trajectory. Stage~3 is distribution matching on self-forced rollouts, with a
small MSE anchor of the supervised video chunk to the ground-truth latent.
Without the anchor the critic absorbs high-frequency noise and stops
penalizing it. For Table~\ref{tab:geom} the two distillation increments are
merged into one update ($\Delta_{cd} \oplus \Delta_{dmd}$).

\end{document}